\documentclass[10pt,twocolumn,letterpaper]{article}

\usepackage[pagenumbers]{cvpr}

\usepackage[dvipsnames]{xcolor}

\usepackage{graphicx}
\usepackage{multirow}
\usepackage{sepnum}
\usepackage[linesnumbered,ruled,vlined]{algorithm2e}

\definecolor{cvprblue}{rgb}{0.21,0.49,0.74}
\usepackage[pagebackref,breaklinks,colorlinks,citecolor=cvprblue]{hyperref}

\title{Fully Spiking Denoising Diffusion Implicit Models}

\author{Ryo Watanabe$^1$, Yusuke Mukuta$^{1,2}$, Tatsuya Harada$^{1,2}$\\
$^1$ The University of Tokyo\\
$^2$ RIKEN\\
{\tt\small \{r-watanabe, mukuta, harada\}@mi.t.u-tokyo.ac.jp}
}

\begin{document}
\maketitle
\begin{abstract}

  Spiking neural networks (SNNs) have garnered considerable attention owing to their ability to run on neuromorphic devices with super-high speeds and remarkable energy efficiencies. SNNs can be used in conventional neural network-based time- and energy-consuming applications. However, research on generative models within SNNs remains limited, despite their advantages.
  In particular, diffusion models are a powerful class of generative models, whose image generation quality surpass that of the other generative models, such as GANs. However, diffusion models are characterized by high computational costs and long inference times owing to their iterative denoising feature.
  Therefore, we propose a novel approach fully spiking denoising diffusion implicit model (FSDDIM) to construct a diffusion model within SNNs and leverage the high speed and low energy consumption features of SNNs via synaptic current learning (SCL).
  SCL fills the gap in that diffusion models use a neural network to estimate real-valued parameters of a predefined probabilistic distribution, whereas SNNs output binary spike trains.
  The SCL enables us to complete the entire generative process of diffusion models exclusively using SNNs.
  We demonstrate that the proposed method outperforms the state-of-the-art fully spiking generative model.

\end{abstract}
\section{Introduction}
\label{sec:intro}

Spiking neural networks (SNNs) are neural networks designed to mimic the signal transmission in the brain~\cite{snn/survey}.
Information in SNNs is represented as binary spike trains that are subsequently transmitted by neurons. SNNs have not only biological plausibility but also enormous computational efficiency. Owing to their potential to operate with extremely low energy consumption and high-speed performance~\cite{neuromorphic/true_north,snn/energy_efficiency} on neuromorphic devices~\cite{neuromorphic/true_north,neuromorphic/loihi,neuromorphic/snn_hardware}, SNNs have garnered significant attention in recent years.
However, the research on image generation models within SNNs remains limited~\cite{snn/fsvae}.

Diffusion models~\cite{diffusion/ddpm,diffusion/ddim} are types of generative models trained to obtain the reverse process of a predefined forward process that iteratively adds noise to the given data. The generative process is achieved by iteratively denoising a given noise sample from a completely random prior distribution. Diffusion models have been actively explored, as it has been reported that they can generate high-quality images, outperforming state-of-the-art image generation models~\cite{diffusion/ddpm,diffusion/beat_gans} such as GANs~\cite{biggan,stylegan2,vq-gan}. However, diffusion models incur high computational costs and long inference times because of the iterative denoising of generative processes. Therefore, SNNs are promising candidates for accelerating the inference of diffusion models by utilizing SNNs capability for high-speed and low-energy-consumption computation.
However, although many studies on diffusion models have been published in the field of conventional artificial neural networks (ANNs)~\cite{diffusion/ddpm,diffusion/ddim,diffusion/beat_gans,diffusion/d3pm,diffusion/imagen,diffusion/latent_diffusion,diffusion/vq_diffusion,diffusion/p2_weighting}, research on diffusion models in SNNs is limited. To the best of our knowledge, only two studies exist for diffusion models that use SNNs~\cite{snn/sddpm,snn/spiking_diffusion}.

\begin{figure}
  \centering
  \begin{subfigure}{\linewidth}
    \includegraphics[width=\linewidth,page=1]{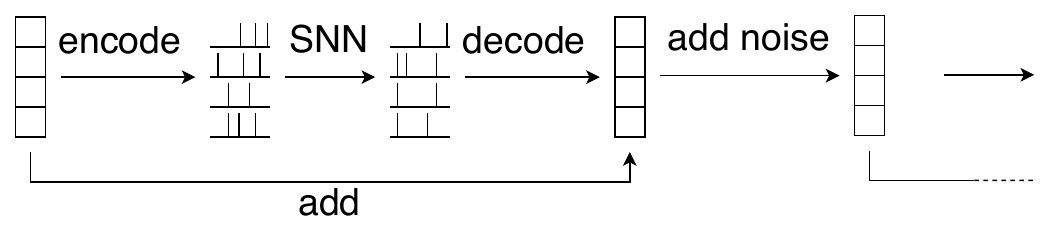}
    \caption{Naively implemented DDPM with an SNN using spikes as input/output. Vertically aligned rectangles represent a vector in the image space. The decoding process requires wait for the final spike emitted.}
    \label{fig:ddpm}
  \end{subfigure}
  \begin{subfigure}{\linewidth}
    \includegraphics[width=\linewidth,page=3]{fig/flow.pdf}
    \caption{Naively implemented DDIM with an SNN using synaptic currents as input/output. DDIM eliminate the need for random noise in each step. The input and output of the SNN are synaptic currents.}
    \label{fig:ddim}
  \end{subfigure}
  \begin{subfigure}{\linewidth}
    \includegraphics[width=\linewidth,page=4]{fig/flow.pdf}
    \caption{DDIM in which the linear combination is computed in signal space via linear encoder and decoder}
    \label{fig:ddim_linear}
  \end{subfigure}
  \begin{subfigure}{0.45\linewidth}
    \includegraphics[width=\linewidth,page=5]{fig/flow.pdf}
    \caption{FSDDIM with an SNN using synaptic currents as input/output}
    \label{fig:fsddim_current}
  \end{subfigure}
  \hfill
  \begin{subfigure}{0.45\linewidth}
    \includegraphics[width=\linewidth,page=6]{fig/flow.pdf}
    \caption{FSDDIM with an SNN using spike trains as input/output}
    \label{fig:fsddim_spike}
  \end{subfigure}
  \caption{Comparison of the generative process of DDPM, DDIM and FSDDIM.}
  \label{fig:comparison}
\end{figure}

The challenge in developing diffusion models for SNNs lies in designing the generation process. For example, DDPM~\cite{diffusion/ddpm}, which is the basis of many recent studies on image-generation diffusion models~\cite{diffusion/latent_diffusion,diffusion/imagen,diffusion/p2_weighting,diffusion/iddpm,diffusion/beat_gans}, uses Gaussian noise in the forward process. Therefore, the generative process is defined as Gaussian.
A neural network is used to estimate the parameters of the Gaussian and a next step denoised image is sampled from it.
However, in the framework of SNNs, the outputs of the spiking neurons are binary spike trains; therefore, we cannot directly estimate the Gaussian parameters, which are real values, using SNNs.
Other diffusion models using different probabilistic distributions, such as the categorical distribution~\cite{diffusion/d3pm,diffusion/vq_diffusion} or gamma distribution~\cite{diffuison/gamma_diffusion} also have the same problem because they use a neural network to estimate the real-valued parameters of the probabilistic distribution. To use an SNN for diffusion models naively, we must decode the output of the SNN to calculate the parameters of the probabilistic distribution and encode the sampled image from the distribution to feed it to the SNN in the next denoising step.
Although these steps are repeated many times, \eg 1000 times, to obtain the final denoised result, the subsequent sampling process after decoding cannot be performed within the framework of SNNs. In addition, this decoding process hinders the full utilization of SNNs because we must wait until the final spike is emitted, whereas the output of the SNN model is subsequently emitted as spike trains over time (\cref{fig:ddpm}).
Existing studies on SNN diffusion have not solved this problem. Both SDDPM~\cite{snn/sddpm} and Spiking-Diffusion~\cite{snn/spiking_diffusion} must decode the output of the SNN model from spike-formed data to the image output and sample from the calculated probabilistic distribution in each denoising step. Therefore, their work cannot be treated as a "fully spiking" generative model which means that it can be completed by SNN alone while they partially use SNNs.

Here, we propose fully spiking denoising diffusion implicit model (FSDDIM) solving the problem (\cref{fig:fsddim_current} and \cref{fig:fsddim_spike}).
In this study, we adopted DDIM~\cite{diffusion/ddim}, which is a deterministic generative process of Gaussian-based diffusion models, to eliminate the need for random sampling in each denoising step.
A single step of denoising in DDIM is represented by a linear combination of the input and output of the neural network. Hence, if we naively implement a DDIM with an SNN, we must decode the output of the SNN to calculate the linear combination, and encode the result again to feed it into the SNN in the next denoising step. This is the same problem as in existing studies.

To solve this problem, we propose a novel technique called synaptic current learning (SCL).
In SCL, we consider synaptic currents as model outputs instead of spike trains~\cite{snn/weighted_spikes,snn/spiking_resnet,snn/tdbn,snn/srsnn} (\cref{fig:ddim}). Synaptic currents are real-valued signals that are the sum of spikes multiplied by the synaptic weight between the presynaptic and postsynaptic neurons.
At the same time, we use linear functions for the encoder and decoder, which map an image to the synaptic currents and vice versa.
This linear encoding and decoding scheme enabled us to compute scalar multiplication in a linear combination of DDIM with synaptic currents, rather than a decoded image (\cref{fig:ddim}).
Finally, we introduce SCL loss. SCL loss ensured that the output of the SNN model could be reconstructed after decoding and encoding. Owing to this loss, we could remove the encoding and decoding processes in each denoising step and complete the entire generative process using the SNN alone (\cref{fig:fsddim_current}). Additionally, we showed that our model, which outputs synaptic currents, can be equivalently regarded as a model that outputs spike trains by fusing the linear combination weights and the first and last convolutional layers into a single convolutional layer (\cref{fig:fsddim_spike}).

To the best of our knowledge, FSDDIM is the first to complete the entire generative process of diffusion models exclusively using SNNs.
Finally, we demonstrate that the proposed method outperforms the state-of-the-art fully spiking image generation model for SNNs.

We summarize our contributions as follows:
\begin{itemize}
  \item We propose a novel approach named Fully Spiking Denoising Diffusion Implicit Models (FSDDIM) to build a diffusion model within SNNs.
  \item We introduce the Synaptic Current Learning (SCL) adopting synaptic currents as model outputs and linear encoding and decoding scheme. We show that the SCL loss enables us to complete the entire generative process of our diffusion model exclusively using SNNs.
  \item We demonstrate that our proposed method outperforms the state-of-the-art fully spiking image generation model in terms of the quality of generated images and the number of time steps.
\end{itemize}

\section{Related work}
\label{sec:related_work}

\subsection{Spiking neural networks}
SNNs model the features of a biological brain, in which the signal is emitted as spike trains, which are time-series binary data. The emitted spikes change the membrane potential of the postsynaptic neuron; when the membrane potential reaches a threshold value, the neuron emits a spike. The membrane potential is reset to the resting potential after the spike emission. We use the iterative leaky integrate-and-fire (LIF) model~\cite{snn/iterative_lif} in this study. The iterative LIF model is the first-order Euler approximation of the LIF model~\cite{snn/lif} and is defined as follows:
\begin{equation}
  u_s = \tau_{decay} u_{s-1} + I_s
\end{equation}
where $u_s$ is the membrane potential at the time step $s = 1,\ldots,S$. $S$ is the number of time steps. $\tau_{decay}$ is the decay constant and $I_s$ is the synaptic current at time  step $s$. The neuron emits a spike when the membrane potential $u_s$ reaches the threshold value $V_{th}$ and the membrane potential is reset to $0$ after emitting the spike. This is expressed as follows:
\begin{equation}
  u^i_{s, n} = \tau_{decay} u^i_{s-1, n} (1 - o^i_{s-1,n}) + I^i_{s, n-1}
  \label{eq:memblane_potential_update}
\end{equation}
\begin{equation}
  o^i_{s, n} =  \begin{cases}
    1 & \text{if } u^i_{s, n} \geq V_{th} \\
    0 & \text{otherwise}
  \end{cases}
  \label{eq:lif_firing}
\end{equation}
Here, $u^i_{s, n}$ and $o^i_{s, n}$ are the membrane potential and the output of the $i$th neuron in the $n$th layer at time step $s$ respectively.

The synaptic current $I^i_{s, n-1}$ is the weighted sum of all spikes transmitted by the neurons of the previous layer and is calculated by
\begin{equation}
  I^i_{s, n-1} = \sum_{j} w_j o^j_{s, n-1}
  \label{eq:pre_synaptic_input}
\end{equation}
$j$ is the index of neurons connected to the neuron and $w_j$ is the synaptic weight of each connection. $o^j_{s, n-1}$ is the output of the $j$th neuron at time step $s$ in the $n-1$th layer.

\subsection{Training of SNNs}

Learning algorithms for SNNs have been actively investigated in recent years. Legenstein \etal~\cite{snn/stdp} used spike timing-dependent plasticity (STDP)~\cite{stdp} to train SNNs. Diehl \etal\cite{snn/stdp2} proposed an unsupervised learning method using STDP and achieved $95\%$ accuracy on MNIST~\cite{mnist}. Recently, backpropagation has been widely used to train SNNs~\cite{snn/backprop1,snn/backprop2,snn/backprop3,snn/backprop4, snn/surrogate_gradient,snn/surrogate_gradient2,snn/surrogate_gradient_arctan} and these methods have achieved high performance compared with other existing methods. As the firing mechanisms in spiking neurons, such as those in \cref{eq:lif_firing}, are non-differentiable, surrogate gradient functions are used to make backpropagation possible. Zenke \etal~\cite{snn/sg_robustness} systematically demonstrated the robustness of surrogate gradient learning.

In this study, we used surrogate gradient learning to train SNNs. We used a surrogate gradient function similar to that in \cite{snn/surrogate_gradient, snn/sg_robustness} which is defined as
\begin{equation}
  \begin{split}
    \frac{\partial o^i_{s, n}}{\partial u^i_{s, n}} = \max\left( 1 - \frac{1}{a} \left|u^i_{s,n} - V_{th}\right|, 0 \right)
  \end{split}
  \label{eq:surrogate_gradient}
\end{equation}

\subsection{Diffusion models}
\label{sec:diffusion_models}
Diffusion models are generative models that generate the original data via iterative denoising. Ho \etal~\cite{diffusion/ddpm} proposed the denoising diffusion probabilistic model (DDPM) and demonstrated the high-quality image generation ability of diffusion models. This work was followed by numerous other reported studies in later years~\cite{diffusion/beat_gans,diffusion/iddpm,diffusion/latent_diffusion,diffusion/imagen,diffusion/ddim,diffusion/vq_diffusion}.  The forward process of DDPM is a Markov chain that gradually adds Gaussian noise.
\begin{equation}
  q(x_t \mid x_{t-1}) = \mathcal{N}(x_t; \sqrt{\alpha_t} x_{t-1}, (1  -\alpha_t) I)
  \label{eq:ddpm}
\end{equation}
$x_0$ is the given data and iteratively adds noise defined in \cref{eq:ddpm} from $t=1$ to $T$ such that $x_T$ follows a standard Gaussian distribution $p(x_T) = \mathcal{N}(x_T; 0, I)$. This definition enables us to sample $x_t$ in closed form using $\bar{\alpha}_t=\prod^t_{s=1}\alpha_s$
\begin{equation}
  q(x_t \mid x_0) = \mathcal{N}(x_t; \sqrt{\bar{\alpha}_t} x_0, (1  -\bar{\alpha}_t) I)
  \label{eq:ddpm_sample}
\end{equation}
While the training objective of reverse process $p_\theta(x_{0:T})$ is derived from the variational bound on negative log likelihood $\mathbb{E}\left[ -\log p_\theta(x_0) \right]$, Ho \etal~\cite{diffusion/ddpm} and later works~\cite{diffusion/p2_weighting,diffusion/min_snr} found that the loss function \cref{eq:ddpm_loss_weighted} improves the performance instead of the variational bound.
\begin{equation}
  L_\gamma= \sum^T_{t=1}
  \gamma_t
  \mathbb{E}\left[\left\| \epsilon - \epsilon^{(t)}_\theta \left(\sqrt{\bar{\alpha}_t} x_0 + \sqrt{1  -\bar{\alpha}_t}\epsilon\right) \right\|^2\right]
  \label{eq:ddpm_loss_weighted}
\end{equation}
where $\epsilon$ is a noise sampled from the standard Gaussian distribution $\mathcal{N}(0,I)$ and $\epsilon^{(t)}_\theta$ is the noise estimated using a neural network.

Song \ et al. ~\cite{diffusion/ddim} studied a non-Markovian definition of diffusion models and showed that the variational inference objective under their definition is equivalent to the weighted loss function of DDPM in \cref{eq:ddpm_loss_weighted}. Remarkably, their generative process allows deterministic sampling, called denoising diffusion implicit model (DDIM):
\begin{multline}
  x_{t-1} = \sqrt{\bar{\alpha}_{t-1}}
  \left( \frac{x_t - \sqrt{1-\bar{\alpha}_t}\epsilon^{(t)}_\theta(x_t)}{\sqrt{\bar{\alpha}_t}} \right) \\
  + \sqrt{1  -\bar{\alpha}_{t-1}}\epsilon^{(t)}_\theta(x_t)
  \label{eq:ddim_sample}
\end{multline}
They also showed that DDIM could generate better-quality images in fewer steps than DDPM.

Other studies used different forms of the target predicted by neural networks, such as directly estimating the original image $x_0$~\cite{diffusion/vq_diffusion,diffusion/velocity} and velocity~\cite{diffusion/velocity}. Because $x_t$ can be sampled from \cref{eq:ddpm_sample}, the relationship between $x_t$, $\epsilon$ and $x_0$ can be written as
\begin{equation}
  x_t = \sqrt{\bar{\alpha}_t} x_0 + \sqrt{1  -\bar{\alpha}_t}\epsilon
  \label{eq:ddpm_reparam}
\end{equation}
Velocity $v_t$ is defined as follows:
\begin{equation}
  v_t = \sqrt{\bar{\alpha}_t}\epsilon - \sqrt{1  -\bar{\alpha}_t} x_0
  \label{eq:velocity}
\end{equation}

Regardless of the targets $\epsilon$, $x_0$ and $v_t$, the denoising step of DDIM is represented as a linear combination of the input and output of the neural network $f_{\mathrm{NN}}(x_t, t)$:
\begin{equation}
  x_{t-1} = a_t x_t + b_t f_{\mathrm{NN}}(x_t, t)
  \label{eq:ddim}
\end{equation}
where $a_t$ and $b_t$ are constants calculated from $\bar{\alpha}_t$ and $\bar{\alpha}_{t-1}$.

\subsection{Spiking diffusion models}

Cao \etal~\cite{snn/sddpm} developed the spiking denoising diffusion probabilistic model (SDDPM), which simply replaced the U-Net~\cite{pixelcnn++,unet} used in DDPM~\cite{diffusion/ddpm} with an SNN called Spiking U-Net to enjoy the low energy consumption of SNNs.
They additionally proposed threshold guidance to efficiently train the Spiking U-Net by adjusting the threshold $V_{th}$.
They reported that SDDPM outperformed other image generation models of SNNs~\cite{snn/fsvae,snn/spiking_diffusion}.
However, it is problematic to treat their work as a "fully spiking" generative model which means that it can be completed by SNN alone.
When calculating a single step of denoising in SDDPM, we first need to encode the input image into spike-formed data that SNNs can accept. We then decode the output of the Spiking U-Net from the spike-formed data to the image output. Finally, we calculate the parameters of the Gaussian distribution and sample the next step input image in the same manner as DDPM. As this calculation is performed after decoding, the following sampling process cannot be performed within the SNN.

Spiking-Diffusion~\cite{snn/spiking_diffusion}, another existing SNN diffusion model, uses discrete diffusion models~\cite{diffusion/d3pm} instead of Gaussian-based diffusion models. To obtain a discrete representation of the dataset that they wanted to reproduce, they proposed a vector-quantized spiking variational encoder.
However, they also calculated the denoising step by decoding the output of the SNN and sampling it from the calculated probabilistic distribution. Therefore, their study could not be completed using SNNs alone.
\section{Method}
\label{sec:method}

SNNs handle spike trains and binary time-sequential data because the outputs of a spiking neuron are binary spike trains. However, many widely studied image-generation-diffusion models based on a Gaussian noise distribution~\cite{diffusion/ddpm,diffusion/iddpm,diffusion/latent_diffusion,diffusion/beat_gans,diffusion/imagen,diffusion/distillation} require a neural network to estimate a real-valued target. Therefore, it is difficult to apply these diffusion models to SNNs.

Our model adopts Gaussian-based diffusion models to make it possible to incorporate knowledge from numerous studies on Gaussian-based diffusion models~\cite{diffusion/iddpm, diffusion/p2_weighting,diffusion/min_snr,diffusion/noise_scheduling,diffusion/beat_gans,diffusion/velocity,diffusion/sigmoid}. At the same time, our model can perform the entire generative process exclusively using SNNs, which is a crucial capability to ensure seamless integration with neuromorphic devices to enjoy the benefits of low energy consumption and high-speed calculation in the future.

\subsection{Model overview}

\begin{figure*}[t]
  \centering
  \includegraphics[width=0.8\linewidth]{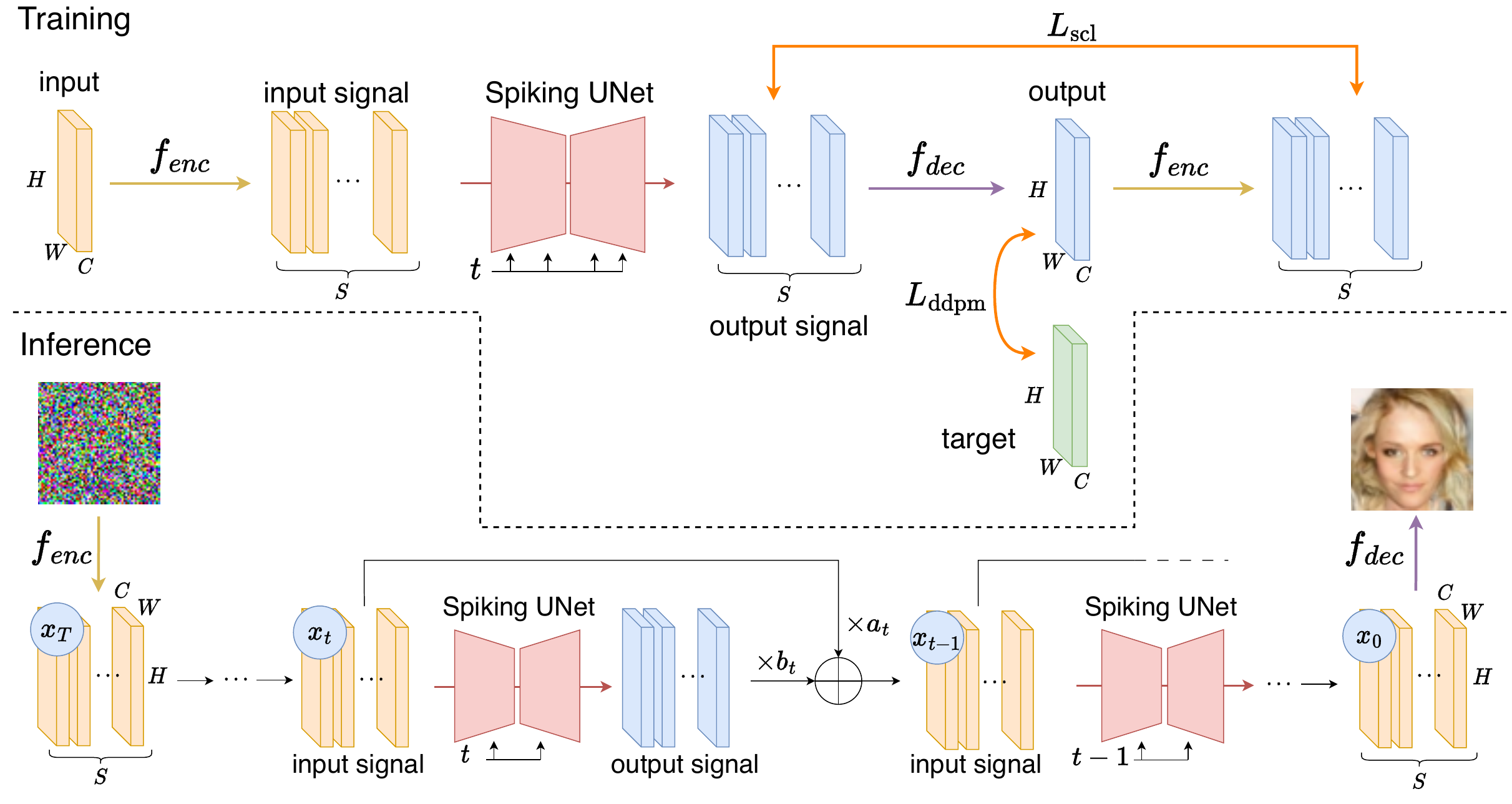}

  \caption{The overview of our proposed method.}
  \label{fig:pipeline}
\end{figure*}

We describe our pipeline in \cref{fig:pipeline}. We use DDIM~\cite{diffusion/ddim} as the basis of our diffusion model. DDIM is a deterministic generative process of Gaussian-based diffusion models. In inference phase, while a single denoising step of DDIM is computed by a linear combination of the input and output of a neural network, we enable our SNN model to calculate it without decoding the output of the SNN to original image space by proposed SCL.

At first, we incorporate an idea to regard synaptic currents as model outputs instead of spike trains. The synaptic currents are real-valued signals calculated by \cref{eq:pre_synaptic_input}. The idea of using synaptic currents as the model output is not new and has been used in previous studies.
Some studies that solved classification problems with SNNs~\cite{snn/weighted_spikes,snn/spiking_resnet} used synaptic currents to calculate the final prediction by accumulating them using the membrane potential of non-firing neurons.
Zheng \etal~\cite{snn/tdbn} also accumulated synaptic currents but averaged them over time.
Cherdo \etal~\cite{snn/srsnn} proposed SRSNN which uses SNN as a recurrent neural network and directly minimizes the synaptic currents and target values.

Next, in the following section, we analyze DDIM naively implemented with an SNN in which the input is encoded to synaptic currents and the output is decoded to the original image space (\cref{fig:ddim}). As a result, we show that the encoding and decoding process can be removed by introducing the SCL loss utilizing linear encoder and decoder.
In this way, the denoising step can be completed by a linear combination of the input and output synaptic currents of the SNN (\cref{fig:fsddim_current}).
This linear combination is processed independently for each time step, like ResNet in SNNs~\cite{snn/tdbn,snn/sew_resnet,snn/ms_resnet,snn/ems_resnet}. Therefore, our diffusion model can process the denoising step without a delay in decoding the output of the SNN or by depending on modules other than SNNs.
Additionally, we show that our model which outputs synaptic currents can be equivalently regarded as a model which outputs spike trains by fusing the linear combination weights and the first and last convolutional layer into a single convolutional layer (\cref{fig:fsddim_spike}).
We call our diffusion model FSDDIM.

\subsection{SCL for SNN diffusion}

In this section, we describe how to train an SNN for FSDDIM and introduce SCL. First, we begin with the regular DDIM~\cite{diffusion/ddim}. A single step in the generative process of DDIM is represented as \cref{eq:ddim}.

Next, we consider replacing the neural network $f_{\mathrm{NN}}$ with an SNN $f_{\mathrm{SNN}}$ but use encoder $f_{\mathrm{enc}}$ and decoder $f_{\mathrm{dec}}$ to convert values in the original image space into time sequence signals that the SNNs take as inputs and outputs, and vice versa.
Because our SNN model outputs synaptic currents, we define the encoder $f_{\mathrm{enc}}(\cdot)$ as a function that maps an image $x\in \mathbb{R}^{H\times W \times C}$, with the shape of $H\times W$ and $C$ channels, to the sequential synaptic current signals $x^\prime\in\mathbb{R}^{H \times W \times C \times S}$.
The decoder $f_{\mathrm{dec}}(\cdot)$ is a function that maps sequential synaptic current signals $x^\prime\in\mathbb{R}^{H \times W \times C \times S}$ to an image $x\in \mathbb{R}^{H\times W \times C}$.
Here, we refer to the original data space $\mathbb{R}^{H\times W \times C}$ as the image space and the synaptic current signal space $\mathbb{R}^{H \times W \times C \times S}$ as the signal space.
The input of the SNN $f_{\mathrm{SNN}}$ is the synaptic current signals $x_t^\prime\in\mathbb{R}^{H \times W \times C \times S}$ and the step $t\in\{1,\ldots,T\}$. The output is the synaptic current signals $\hat{z}^{\prime}\in\mathbb{R}^{H \times W \times C \times S}$.

Using SNN $f_{\mathrm{SNN}}$, encoder $f_{\mathrm{enc}}$ and decoder $f_{\mathrm{dec}}$, the denoising step becomes
\begin{equation}
  x_{t-1} = a_t x_t + b_t f_{\mathrm{dec}}\left({f_{\mathrm{SNN}}(f_{\mathrm{enc}}(x_t), t)}\right)
  \label{eq:ddim_enc_dec}
\end{equation}
(\cref{fig:ddim}).
The loss was calculated in the image space using \cref{eq:ddpm_loss_weighted}.
\begin{equation}
  L_\mathrm{ddpm}= \sum^T_{t=1}
  \gamma_t
  \mathbb{E}\left[\bigl\| z -f_{\mathrm{dec}}\left({f_{\mathrm{SNN}}(f_{\mathrm{enc}}(x_t),t)}\right) \bigr\|^2\right]
  \label{eq:ddpm_loss_with_snn}
\end{equation}
$z$ is the target such as the noise $\epsilon$, original image $x_0$, or velocity $v_t$, which the SNN tries to estimate.
Subsequently, we obtained a generative process using an SNN. However, it still requires encoding and decoding at each step of the generative process and cannot be completed by only SNNs. To solve this problem, we aim to calculate the generative process of DDIM in the signal space (\cref{fig:fsddim_current}).
\begin{equation}
  x_{t-1}^\prime = a_t x_t^\prime + b_t {f_{\mathrm{SNN}}(x_t^\prime, t)}
  \label{eq:ddim_signal}
\end{equation}
where $x_t^\prime$ and $x_{t-1}^\prime$ are corresponding signals to $x_t$ and $x_{t-1}$ in the signal space.

First, we propose the use of a linear encoder and decoder for $f_{\mathrm{enc}}$ and $f_{\mathrm{dec}}$ to compute the linear combination in the signal space (\cref{fig:ddim_linear}). Because the encoder and decoder are linear maps, we can change the order of these operations and the linear combination. Therefore, we can rewrite \cref{eq:ddim_enc_dec} as follows:
\begin{equation}
  x^\prime_{t-1} = a_t x^\prime_t + b_t f_{\mathrm{enc}}\left(f_{\mathrm{dec}}\left({f_{\mathrm{SNN}}(x^\prime_t, t)}\right)\right)
  \label{eq:ddim_linear}
\end{equation}
where $x^\prime_T=f_{\mathrm{enc}}(x_T)$.

Then, we can remove the decoding and encoding processes in \cref{eq:ddim_linear} by approximating $f_{\mathrm{enc}}(f_{\mathrm{dec}}(f_{\mathrm{SNN}}(x^\prime_t, t)))$ with $f_{\mathrm{SNN}}(x^\prime_t, t)$ by minimizing the squared error $L_{\mathrm{scl}}$ between them at every step.
\begin{multline}
  L_{\mathrm{scl}} = \sum_{t=1}^{T} \lambda_t \\\mathbb{E} \left[ \bigl\|f_{\mathrm{SNN}}(x^\prime_t, t) - f_{\mathrm{enc}}
  \left(f_{\mathrm{dec}}\left( f_{\mathrm{SNN}}(x^\prime_t, t) \right)\right)  \bigr\|^2 \right]
  \label{eq:scl_single}
\end{multline}
$\lambda_t$ is a hyper-parameter. Consequently, we obtain \cref{eq:ddim_signal}.

We call the loss function $L_{\mathrm{scl}}$ the SCL loss and simultaneously minimize the SCL and diffusion losses \cref{eq:ddpm_loss_with_snn}. The total loss function $L$ becomes:
\begin{equation}
  L = L_{\mathrm{ddpm}} + L_{\mathrm{scl}}
  \label{eq:loss_total}
\end{equation}

\subsection{Equivalence between synaptic currents and spike trains}

\begin{figure*}[t]
  \centering
  \includegraphics[width=0.8\linewidth]{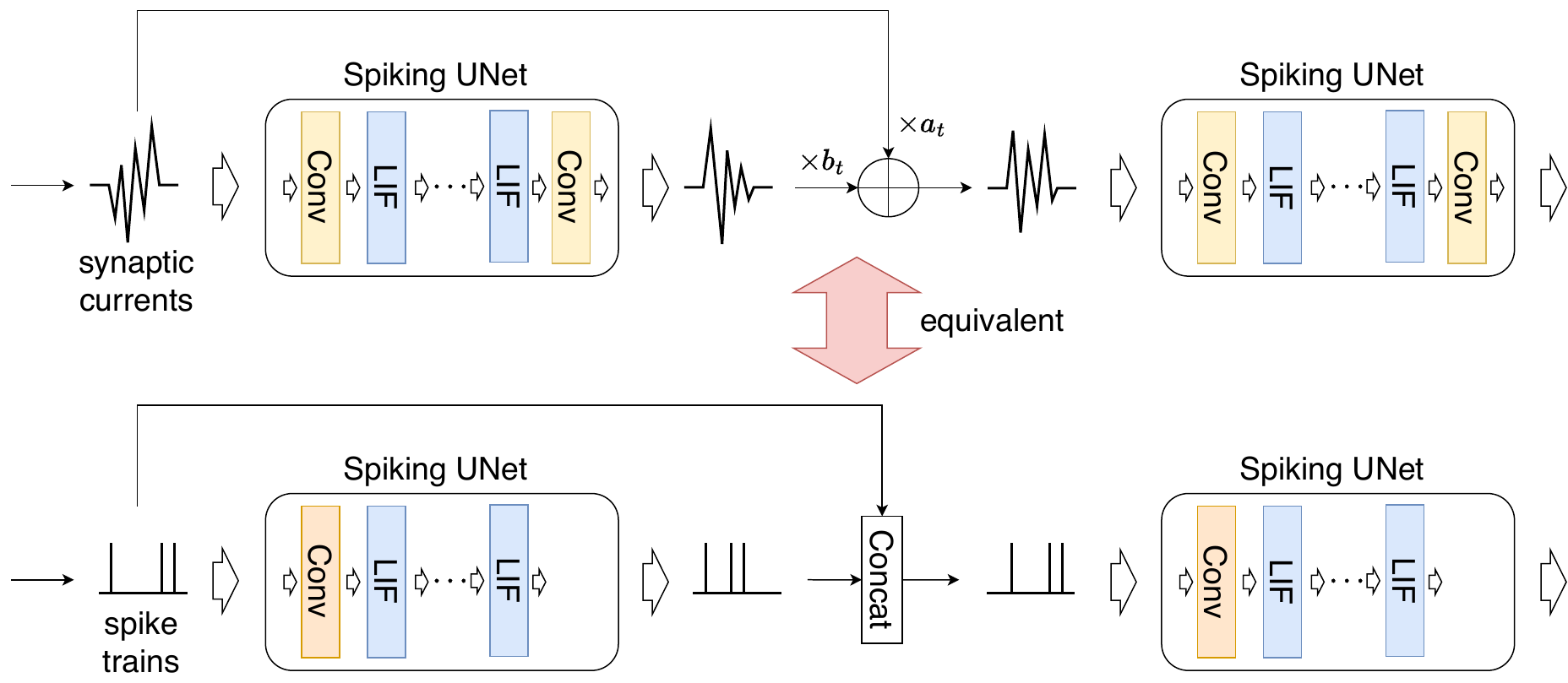}

  \caption{The conversion from the linear combination of synaptic currents to the channel-wise concatenation of spike trains. We can regard our model which outputs synaptic currents as a model which outputs spike trains by fusing the linear combination weights and the first and last convolutional layer into a single convolutional layer.}
  \label{fig:current2spike}
\end{figure*}

Our denoising step involves the linear combination and subsequent convolution of real-valued synaptic currents. While recent works for SNNs accept non-binary multiply-add operations~\cite{snn/sew_resnet, snn/ms_resnet, snn/spikformer, snn/attention}, our model can be further equivalently regarded as processing only spike trains, as illustrated in \cref{fig:current2spike}. The linear combination, as well as the first and last convolutions in the SNN, constitute affine maps and thus can be consolidated into a single convolutional layer.

\subsection{Network architecture}

We designed our SNN architecture based on the UNet used in DDPM~\cite{diffusion/ddpm} without self-attention. Instead of group normalization~\cite{group_norm}, we use tdBN~\cite{snn/tdbn}.
We also added the time embeddings for step $t$ to each ResNet block in our Spiking UNet following DDPM. We used direct input encoding~\cite{snn/direct_input} for time embeddings.

In the following section, we use direct input encoding~\cite{snn/direct_input} calculated independently of each pixel for encoder $f_{\mathrm{enc}}$, which is defined as
\begin{equation}
  f_{\mathrm{enc}}(x_{ijk}) = (x_{ijk}, x_{ijk}, \ldots, x_{ijk})
  \label{eq:direct_input}
\end{equation}
where $x_{ijk}$ denotes the pixel value of the input image at position $(i, j)$ and channel $k$. The decoder $f_{\mathrm{dec}}$ is defined similarly to \cite{snn/tdbn} as
\begin{equation}
  f_{\mathrm{dec}}\left(x^{(1:S)}_{ijk}\right) = \frac{1}{S} \sum_{s=1}^S x_{ijk}^{(s)}
  \label{eq:average_output}
\end{equation}
where $x^{(s)}_{ijk}$ is the output synaptic current of the index $(i,j,k)$ at the time step $s$.

\section{Experiment}
\label{sec:experiment}

We implemented our model using PyTorch~\cite{pytorch} and trained it using MNIST~\cite{mnist}, Fashion MNIST~\cite{fashion_mnist}, CIFAR10~\cite{cifar10} and CelebA~\cite{celeba}. Because there are no other fully spiking diffusion models, we compared the evaluation metrics to FSVAE~\cite{snn/fsvae}, the state-of-the-art VAE-based fully spiking generative model for SNNs.

\subsection{Experimental settings}

The thresholds $V_{th}$ and $\tau_{decay}$ were set to $1.0$ and $0.8$ respectively. During backpropagation, we use a detaching reset~\cite{snn/sg_robustness}. We used $4$ or $8$ as the number of time steps $S$. Generally, higher time steps have more expressive power; however, they also increase the computational cost and latency.

The diffusion model uses a cosine schedule~\cite{diffusion/iddpm,diffusion/velocity} with the number of steps $T=1000$, and our Spiking UNet is trained to predict the velocity $v_t$ following \cite{diffusion/velocity}.

We used the Adam optimizer~\cite{adam} with $\beta_1=0.9$, $\beta_2=0.999$ and $\epsilon=10^{-8}$ with a learning rate of $0.001$ for MNIST and Fashion MNIST, and $0.0005$ for CIFAR10 and CelebA. The batch size is $256$.

For evaluation metrics of the quality of generated images, we used the same metrics as FSVAE~\cite{snn/fsvae}, Fr\'echet inception distance (FID)~\cite{fid} and Fr\'echet autoencoder distance (FAD). FAD is a metric used in FSVAE and is calculated by the Fr\'echet distance of a trained autoencoder's latent variables between samples and real images. They suggested the use of FAD because FID uses the output of Inception-v3~\cite{inception_v3} trained on ImageNet~\cite{imagenet} and it may not work in a different domain such as MNIST. We used the same autoencoder architecture as FSVAE for FAD. Following FSVAE, we used the test data of each dataset for evaluation: \sepnum{.}{,}{,}{10000} images from MNIST, Fashion MNIST and CIFAR10, and \sepnum{.}{,}{,}{19962} images from CelebA. We then generate \sepnum{.}{,}{,}{10000} images for each dataset and the metrics were calculated.

\subsection{Results}

The evaluation results are presented in \cref{tab:results}.
We can see that FSDDIM outperforms FSVAE~\cite{snn/fsvae} in terms of FID and FAD for all datasets. Moreover, FSDDIM achieved much higher results with fewer time steps than FSVAE, which reduced the computational cost and latency.
We successfully trained the model with $55.3$M parameters, whereas Kamata \etal investigated relatively small models with a number of parameters up to $6.7$M for FSVAE~\cite{snn/fsvae}.
These results suggest that the superiority of diffusion models over VAEs is also valid in the case of SNNs.

In addition, we compared the FID with the existing SDDPM~\cite{snn/sddpm} and Spiking-Diffusion ~\cite{snn/spiking_diffusion} which are not fully spiking but diffusion models using SNNs. As a result, FSDDIM achieved better FID than that of both models on MNIST and Fashion MNIST, and than that of Spiking-Diffusion on CIFAR10. Therefore, FSDDIM is superior to these existing works not only in terms of the potential of the model to be seamlessly implemented on neuromorphic devices in the future, but also in terms of the quality of the generated images.

\begin{figure*}
  \centering
  \begin{subfigure}{0.22\linewidth}
    \includegraphics[width=\linewidth]{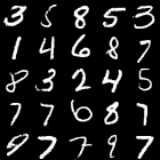}
    \caption{MNIST}
    \label{fig:mnist}
  \end{subfigure}
  \hfill
  \begin{subfigure}{0.22\linewidth}
    \includegraphics[width=\linewidth]{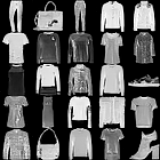}
    \caption{Fashion MNIST}
    \label{fig:fashion}
  \end{subfigure}
  \hfill
  \begin{subfigure}{0.22\linewidth}
    \includegraphics[width=\linewidth]{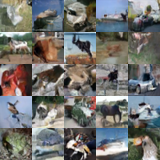}
    \caption{CIFAR10}
    \label{fig:cifar10}
  \end{subfigure}
  \hfill
  \begin{subfigure}{0.22\linewidth}
    \includegraphics[width=\linewidth]{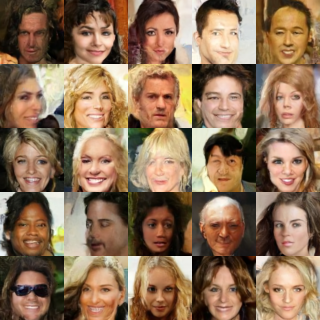}
    \caption{CelebA}
    \label{fig:celeba}
  \end{subfigure}
  \caption{Generated images of MNIST, Fashion MNIST, CIFAR10 and CelebA. For CelebA, relatively good results are selected. The results on the other datasets are randomly sampled. The results of Fashion MNIST and CelebA are generated with $4$ time steps. The results of MNIST and CIFAR10 are generated with $8$ time steps.}
  \label{fig:generated_samples}
\end{figure*}

The images generated by FSVAE are shown in \cref{fig:generated_samples}.

\begin{table*}
  \centering
  \begin{tabular}{ccccc}
    \toprule
    Dataset                        & Model                                          & Time steps & FID $\downarrow$ & FAD $\downarrow$ \\
    \midrule
    \multirow{5}{*}{MNIST}         & SDDPM~\cite{snn/sddpm}                         & 4          & $(29.48)$        & $-$              \\
                                   & Spiking-Diffusion~\cite{snn/spiking_diffusion} & 16         & $(37.50)$        & $-$              \\
                                   & FSVAE~\cite{snn/fsvae}                         & 16         & $57.77$          & $36.52$          \\
                                   & \textbf{FSDDIM (ours)}                         & 8          & $\mathbf{3.99}$  & $5.71$           \\
                                   & \textbf{FSDDIM (ours)}                         & 4          & $7.48$           & $\mathbf{3.62}$  \\
    \midrule
    \multirow{5}{*}{Fashion MNIST} & SDDPM~\cite{snn/sddpm}                         & 4          & $(21.38)$        & $-$              \\
                                   & Spiking-Diffusion~\cite{snn/spiking_diffusion} & 16         & $(91.98)$        & $-$              \\
                                   & FSVAE~\cite{snn/fsvae}                         & 16         & $58.45$          & $16.45$          \\
                                   & \textbf{FSDDIM (ours)}                         & 8          & $11.78$          & $\mathbf{4.91}$  \\
                                   & \textbf{FSDDIM (ours)}                         & 4          & $\mathbf{9.17}$  & $9.25$           \\
    \midrule
    \multirow{6}{*}{CIFAR10}       & SDDPM~\cite{snn/sddpm}                         & 8          & $(16.89)$        & $-$              \\
                                   & SDDPM~\cite{snn/sddpm}                         & 4          & $(19.20)$        & $-$              \\
                                   & Spiking-Diffusion~\cite{snn/spiking_diffusion} & 16         & $(120.5)$        & $-$              \\
                                   & FSVAE~\cite{snn/fsvae}                         & 16         & $175.7$          & $139.97$         \\
                                   & \textbf{FSDDIM (ours)}                         & 8          & $\mathbf{46.14}$ & $12.61$          \\
                                   & \textbf{FSDDIM (ours)}                         & 4          & $51.46$          & $\mathbf{8.63}$  \\
    \midrule
    \multirow{3}{*}{CelebA}        & SDDPM~\cite{snn/sddpm}                         & 4          & $(25.09)$        & $-$              \\
                                   & FSVAE~\cite{snn/fsvae}                         & 16         & $101.5$          & $113.15$         \\
                                   & \textbf{FSDDIM (ours)}                         & 4          & $\mathbf{36.08}$ & $\mathbf{66.52}$ \\
    \bottomrule
  \end{tabular}
  \caption{Comparison of FID and FAD on MNIST, Fashion MNIST, CIFAR10 and CelebA. Since SDDPM~\cite{snn/sddpm} and Spiking-Diffusion~\cite{snn/spiking_diffusion} are not fully spiking models, we show their results for only reference purpose, and we put the results in parentheses. Their results are taken from \cite{snn/sddpm} and \cite{snn/spiking_diffusion} respectively. The best results among fully spiking models are highlighted in bold.}
  \label{tab:results}
\end{table*}

\subsection{Ablation study}

An ablation study is conducted to investigate the effects of the proposed SCL loss method. We trained our model with and without SCL loss on CIFAR10 and calculated FID with generated \sepnum{.}{,}{,}{5000} images. We used time steps $S=8$.
In addition, we trained our model with another loss function $L_\mathrm{signal}$ instead of the SCL loss, which computes squared error of encoded target $f_{\mathrm{enc}}(z)$ and output of the SNN:
\begin{equation}
  L_\mathrm{signal} = \sum_{t=1}^T \lambda_t \left\| f_{\mathrm{enc}}(z) - f_{\mathrm{SNN}}(x^\prime_t, t)\right\|^2
  \label{eq:signal_loss}
\end{equation}
If the DDPM loss $L_\mathrm{ddpm}$ in \cref{eq:ddpm_loss_with_snn} is $0$, then $L_{\mathrm{signal}}$ is equivalent to the SCL loss. Therefore, we compared the results of the model trained with $L_{\mathrm{signal}}$ or the SCL loss to investigate the effects of the SCL loss.

\begin{table}[h]
  \centering
  \begin{tabular}{@{}ccc@{}}
    \toprule
    Model                                               & FID $\downarrow$ \\
    \midrule
    FSDDIM w/o $L_\mathrm{scl}$                         & $80.03$          \\
    FSDDIM w/ $L_\mathrm{scl}$                          & $\mathbf{48.75}$ \\
    FSDDIM w/o $L_\mathrm{scl}$, w/ $L_\mathrm{signal}$ & $50.33$          \\
    FSDDIM w/ $L_\mathrm{scl}$, w/ $L_\mathrm{signal}$  & $52.77$          \\
    \bottomrule
  \end{tabular}
  \caption{Ablation study of the SCL loss on CIFAR10.}
  \label{tab:ablation}
\end{table}

The results are presented in \cref{tab:ablation}.
We can see that the model trained with SCL loss achieved the highest performance; $48.75$.
The model trained without SCL loss achieved the worst performance; FID was $80.03$. $L_{\mathrm{signal}}$ also improved the performance but the SCL loss was more effective than $L_{\mathrm{signal}}$.
Consequently, we can conclude that the SCL loss is effective for training our model.

\subsection{Computational cost}

We implemented an ANN model with the same architecture as our Spiking UNet and calculated the number of additions and multiplications during the generation of an image to demonstrate the computational effectiveness of FSDDIM. The results are shown in \cref{tab:mul_add}. We can see that our model with $4$ time steps reduces the number of multiplications by $74.1\%$ and the number of additions by $14.9\%$ compared with the ANN model.
Our model with $8$ time steps increased the number of additions, but still reduced the number of multiplications, which is generally more computationally expensive, by $48.1\%$ compared with the ANN model.
Therefore, our model exhibits computational efficiency compared with the ANN model, with the prospect of significant speedup and energy efficiency on neuromorphic devices. We leave the implementation on neuromorphic devices as future work.

\begin{table}[h]
  \centering
  \begin{tabular}{@{}cccc@{}}
    \toprule
    Model         & Time steps & \#Addition                   & \#Multiplication             \\
    \midrule
    ANN           & $-$        & $4.98\times10^{12}$          & $4.97\times10^{12}$          \\
    \textbf{ours} & 8          & $7.14\times10^{12}$          & $2.58\times10^{12}$          \\
    \textbf{ours} & 4          & $\mathbf{4.24\times10^{12}}$ & $\mathbf{1.29\times10^{12}}$ \\
    \bottomrule
  \end{tabular}
  \caption{Computational cost of a single step of generative process on Fashion MNIST. The results are averaged over $100$ images.}
  \label{tab:mul_add}
\end{table}

\section{Conclusion}
\label{sec:conclusion}

In this work, we proposed the FSDDIM which is a fully spiking diffusion model. This model was implemented via SCL, which uses synaptic currents instead of spike trains as model outputs. We introduced SCL loss by adopting a linear encoding and decoding scheme and showed that this SCL loss aids in completing the entire generative process of our diffusion model exclusively using SNNs.
The experimental results showed that our model outperformed the state-of-the-art fully spiked image generation model in terms of quality of the generated images and number of time steps. To the best of our knowledge, our proposed FSDDIM is the first ever reported fully spiking diffusion model.

\section{Acknowledgement}

This work was partially supported by JST Moonshot R\&D Grant Number JPMJPS2011, CREST Grant Number JPMJCR2015 and Basic Research Grant (Super AI) of Institute for AI and Beyond of the University of Tokyo.

We would like to thank Yusuke Mori and Atsuhiro Noguchi for helpful discussions.
We would also like to thank Lin Gu and Akio Hayakawa for their helpful comments on the manuscript.

\appendix
{
\centering
\large
\vspace{\baselineskip}
\textbf{Supplementary Material}\\
}
\section{Implementation details}
\label{sec:implementation_details}

\subsection{Model architecture}

\begin{figure}[h]
  \centering
  \includegraphics[width=\linewidth]{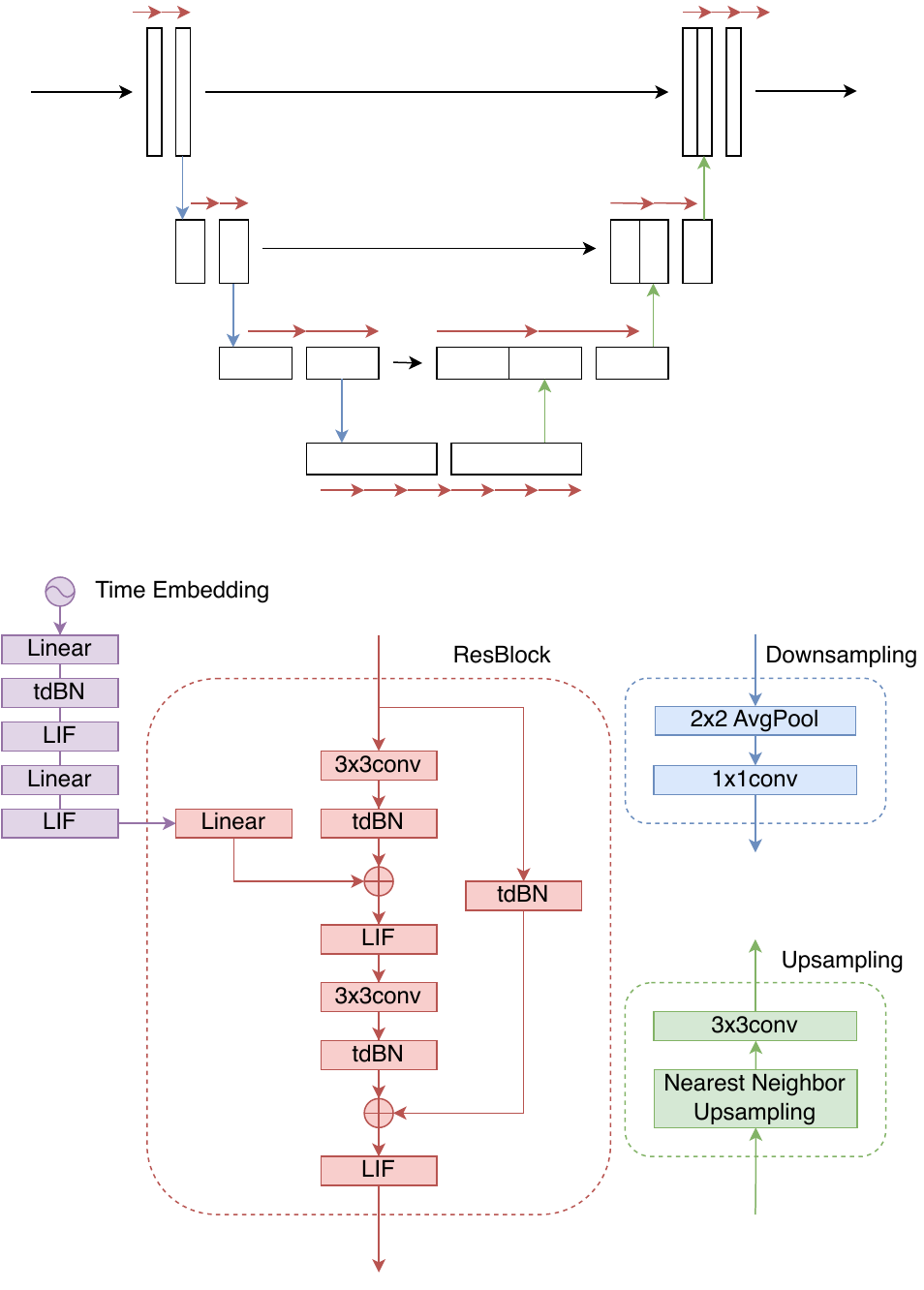}

  \caption{Model architecture.}
  \label{fig:model_architecture}
\end{figure}

Following DDPM~\cite{diffusion/ddpm}, we employ a UNet~\cite{unet,pixelcnn++} as the neural network to estimate the target in each denoising step. The architecture of our Spiking UNet is illustrated in \cref{fig:model_architecture}.
The Spiking UNet comprises four downsampling blocks and four upsampling blocks. Each block consists of two ResBlocks and a downsampling or upsampling layer, except for the last downsampling and upsampling blocks, which contain two ResBlocks and a convolutional layer with tdBN~\cite{snn/tdbn} and LIF layers.
We incorporate time embeddings into the ResBlocks, similar to DDPM. Using sinusoidal time embeddings with 512 dimensions, these embeddings are introduced via direct input encoding~\cite{snn/direct_input} into two linear layers with tdBN and LIF layers. The embeddings are then projected and added in each ResBlock.
The number of channels in each downsampling block is 128, 256, 384, and 512, respectively. Conversely, the number of channels in each upsampling block is 512, 384, 256, and 128. Before the initial downsampling block, a convolutional layer with a channel size of 128 is added. After the final upsampling block, a ResBlock with a channel size of 128 and a convolutional layer with a channel size of 1 or 3 (depending on the dataset) is added.
The convolutional layer in each ResBlock has a kernel size of $3 \times 3$, a stride of $1 \times 1$, and a padding of $1 \times 1$. The downsampling layer consists of an average pooling layer with a kernel size of $2 \times 2$ and a stride of $2 \times 2$, followed by a convolutional layer with a kernel size of $1 \times 1$.
In the upsampling blocks, we utilize a nearest neighbor upsampling layer with a scale factor of $2 \times 2$ and a convolutional layer with a kernel size of $3 \times 3$. The input of each ResBlock is concatenated with the output of the corresponding downsampling block, following the DDPM approach.
The input of the final convolutional layer is concatenated with the output of all ResBlocks with the same resolution. In the inference phase, the subsequent convolutional layers after downsampling and upsampling layers can be fused into a single convolutional layer.

\subsection{Training details}

The images from MNIST, Fashion MNIST, and CIFAR10 were resized to $32 \times 32$. CelebA images were cropped to $148 \times 148$ and then resized to $64 \times 64$. We utilized \sepnum{.}{,}{,}{60000} images for training from MNIST and Fashion MNIST, \sepnum{.}{,}{,}{50000} images from CIFAR10, and \sepnum{.}{,}{,}{162770} images from CelebA.
For MNIST, Fashion MNIST, and CIFAR10, the entire training set was used per epoch. In the case of CelebA, \sepnum{.}{,}{,}{51200} images were randomly selected from the training set for each epoch.
The model was trained for \sepnum{.}{,}{,}{1500} epochs, with FID calculated every 150 epochs using \sepnum{.}{,}{,}{1000} generated images. The best model, determined by FID, was selected for the final evaluation.
We set the hyper-parameters $\gamma_t$ in \cref{eq:ddpm_loss_with_snn} to $\gamma_t=\bar{\alpha}_t$ for all $t$, following the same weighting as DDPM~\cite{diffusion/ddpm} when the target $z$ is the velocity $v_t$. The hyper-parameters $\lambda_t$ in \cref{eq:ddpm_loss_with_snn} were set to $\lambda_t=b_t/S$ for all $t$.
We employed Monte Carlo method to minimize the loss in both \cref{eq:ddpm_loss_with_snn} and \cref{eq:scl_single}, following the approach of DDPM. The training algorithm is presented in \cref{alg:training_algorithm}.

\begin{algorithm}
  \caption{Training Algorithm}
  \label{alg:training_algorithm}

  \Repeat{converged}{
    $x_0\sim q(x_0)$ \\
    $t\sim \mathrm{Uniform}(\{1,\ldots,T\})$ \\
    $\epsilon\sim \mathcal{N}(0, I)$ \\
    $x_t \leftarrow \sqrt{\bar{\alpha}_t} x_0 + \sqrt{1  -\bar{\alpha}_t}\epsilon$ \\
    $v_t \leftarrow \sqrt{\bar{\alpha}_t}\epsilon - \sqrt{1  -\bar{\alpha}_t} x_0$ \\
    $\hat{v} \leftarrow f_\mathrm{SNN}(f_\mathrm{enc}(x_t), t)$ \\
    $l_\mathrm{ddpm} \leftarrow \bigl\| v_t - f_\mathrm{dec}(\hat{v}) \bigr\|^2$ \\
    $l_\mathrm{scl} \leftarrow \bigl\| \hat{v} - f_\mathrm{enc}\left(f_\mathrm{dec}\left(\hat{v}\right)\right) \bigr\|^2$ \\
    Take gradient descent step on
    \[
      \nabla_\theta ( \gamma_t l_\mathrm{ddpm}  + \lambda_t l_\mathrm{scl} )
    \]
  }

\end{algorithm}

\section{Generated images}

\begin{figure}
  \centering
  \includegraphics[width=\linewidth]{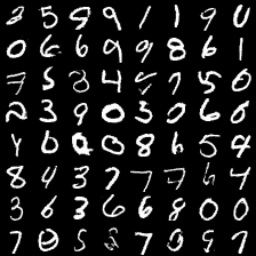}

  \caption{Generated images on MNIST.}
  \label{fig:mnist_additional}
\end{figure}

\begin{figure}
  \centering
  \includegraphics[width=\linewidth]{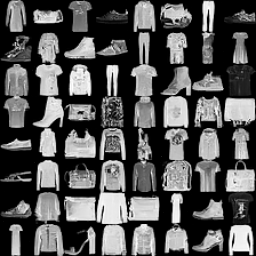}

  \caption{Generated images on Fashion MNIST.}
  \label{fig:fashion_additional}
\end{figure}

\begin{figure}
  \centering
  \includegraphics[width=\linewidth]{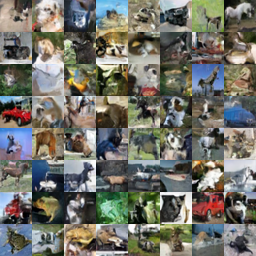}

  \caption{Generated images on CIFAR10.}
  \label{fig:cifar10_additional}
\end{figure}

\begin{figure}
  \centering
  \includegraphics[width=\linewidth]{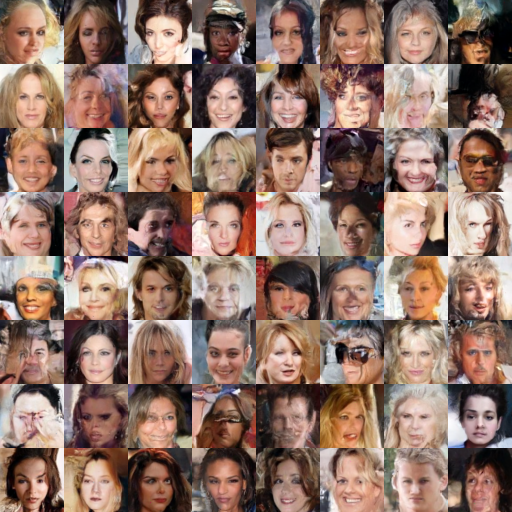}

  \caption{Generated images on CelebA.}
  \label{fig:celeba_additional}
\end{figure}

We show additional generated images of our model in \cref{fig:mnist_additional}, \cref{fig:fashion_additional}, \cref{fig:cifar10_additional} and \cref{fig:celeba_additional}.

\end{document}